\documentclass[letterpaper, 10 pt, conference]{ieeeconf}  % Comment this line out if you need a4paper

\IEEEoverridecommandlockouts                              % This command is only needed if 
\usepackage{graphicx} % modern version of graphics
\usepackage{amsmath} % assumes amsmath package installed
\usepackage{amssymb}  % assumes amsmath package installed
\usepackage{amsfonts}  % assumes amsmath package installed

\usepackage[linesnumbered]{algorithm2e}
\RestyleAlgo{ruled}
\newcommand{\method}{EgoSpeedUp\xspace}
\newcommand*{\sref}[1]{\S\ref{#1}}            % section
\newcommand*{\fref}[1]{\text{Fig.~\ref{#1}}}
\usepackage[nobreak]{cite}

\usepackage[table]{xcolor}
\usepackage{multirow}
\usepackage{booktabs}

\usepackage{caption}
\usepackage{subcaption}

\usepackage{hyperref}
\title{\LARGE \bf
EgoSpeedUp: \\ Transferring Human Manipulation Tempo to Robot Policies}

\author{Hanbit Oh$^{\dagger}$, Yukiyasu Domae, and Takuma Yagi%
\thanks{$^{\dagger}$ Corresponding author}%
\thanks{$^{1}$ The authors are affiliated with the Artificial Intelligence Research Center, National Institute of Advanced Industrial Science and Technology (AIST), Japan.
{\tt\small \{oh.hanbit.oe9, domae.yukiyasu, takuma.yagi\}@aist.go.jp}}%
}

\begin{document}

\maketitle
\thispagestyle{empty}
\pagestyle{empty}

%%%%%%%%%%%%%%%%%%%%%%%%%%%%%%%%%%%%%%%%%%%%%%%%%%%%%%%%%%%%%%%%%%%%%%%%%%%%%%%%
\begin{abstract}
Robot manipulation policies trained through imitation learning inherit not only the demonstrated behavior but also the conservative execution tempo of robot demonstrations. Existing acceleration approaches can execute faster than the original demonstrations, but determine the appropriate acceleration primarily from robot-side information or a predefined set of tempo factors, leaving open how to obtain a task-appropriate reference for how fast each manipulation phase should progress. We introduce \method, a framework that uses human manipulation as temporal supervision for robot imitation learning. Our key insight is that human demonstrations naturally reveal task-appropriate, phase-wise manipulation tempo. Given slow robot demonstrations and human demonstrations of the same task, \method aligns corresponding manipulation phases, estimates their relative execution tempos from multiple human demonstrations, and transfers the resulting phase-wise tempo by retiming the robot demonstrations. The retimed demonstrations are then used for standard behavior cloning, allowing the robot to retain its executable manipulation behavior while learning to perform it at a human-informed tempo. Across two real-world manipulation tasks, \method improves the task success rate by an average of $25$ percentage points (pp) while reducing successful execution time by $36.5\%$. These results demonstrate that human manipulation tempo provides an effective temporal reference for learning faster and more reliable robot policies.
\end{abstract}

%%%%%%%%%%%%%%%%%%%%%%%%%%%%%%%%%%%%%%%%%%%%%%%%%%%%%%%%%%%%%%%%%%%%%%%%%%%%%%%%
\section{Introduction}

Imitation learning has enabled policies to acquire complex robot
manipulation skills from demonstrations
~\cite{zhao2023actionchunking,chi2025diffusionpolicy,yan2025maniflow}.
However, these policies tend to inherit not only the demonstrated behavior but
also its execution tempo. Robot demonstrations are often collected slowly and
conservatively because teleoperation prioritizes precise and successful task
execution over speed~\cite{guo2025demospeedup,nam2025speedaug,kim2026espada}.
Even experienced operators can exhibit less smooth motion during teleoperation than during direct manipulation by hand due to the practical constraints of teleoperation
~\cite{guo2025demospeedup}.
Consequently, even when the same manipulation could be completed much faster,
behavior cloning from slow robot demonstrations tends to reproduce the same
conservative tempo.

Recent studies have shown that robot policies can execute faster than their
demonstrations while retaining task success
~\cite{guo2025demospeedup,arachchige2025sail,kim2026espada,hu2026autospeed,nam2025speedaug}.
Existing approaches achieve this through various forms of temporal adaptation
based primarily on robot demonstrations, robot-side learning objectives, or
subsequent interaction.
Despite their effectiveness, existing approaches determine the target tempo factor from robot-side information, often under manually selected acceleration factors or tempo ranges.
Since an appropriate tempo can vary substantially across tasks, selecting it from such information alone is not straightforward.

\begin{figure}[t]
    \centering
    \includegraphics[width=\linewidth]{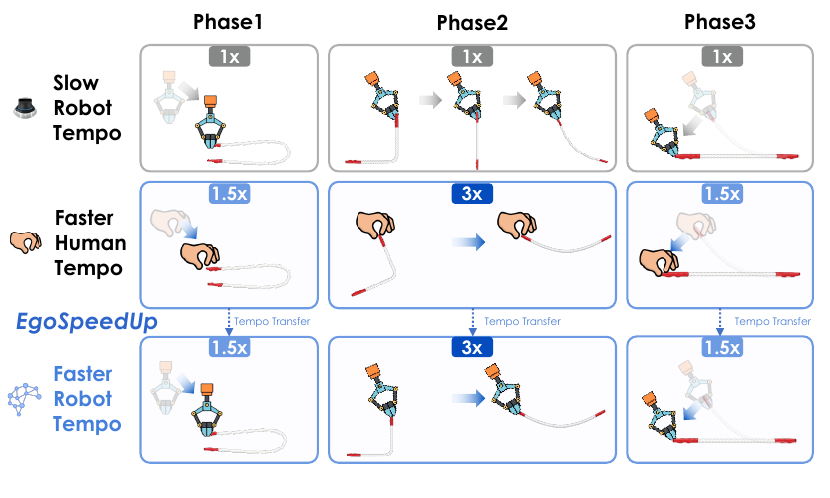}
    \caption{EgoSpeedUp transfers phase-wise manipulation tempo from human to robot demonstrations. Human demonstrations implicitly encode task-appropriate manipulation tempo. For example, humans can efficiently straighten a rope by rapidly swinging it during the motion. EgoSpeedUp aligns slow robot demonstrations with this human tempo, enabling faster robot manipulation while maintaining task success.}
    \label{fig:idea}
\end{figure}

Our key insight is that the timing observed in successful human manipulation
may provide a useful temporal reference for robot manipulation. Human movement
timing has been shown to vary with precision and control requirements
~\cite{wickelgren1977speed}.
For example, in our rope-straightening task, human demonstrations exhibit a
rapid swinging motion, as illustrated in \fref{fig:idea}.
Motivated by this finding, we use the timing of human demonstrations as
\emph{temporal supervision} for estimating the tempo of each manipulation phase.

We propose \method, which transfers this human-derived temporal supervision to robot policies. 
Given slow robot demonstrations and human demonstrations of the same task,
\method aligns corresponding manipulation phases and compares their phase lengths. For each robot phase, phase-length ratios from multiple human
demonstrations are aggregated into a target tempo factor.
These phase-wise tempos
guide the resampling of future robot actions, which are paired with the
original robot observations for behavior cloning. The policy is thereby
trained to perform the demonstrated manipulation at a human-informed tempo,
with the aim of improving execution speed while retaining task success.
Because the target tempos are derived directly from the corresponding human
demonstrations, \method does not require manually selected acceleration
factors or tempo ranges for each new task or object.

We evaluate \method on two real-world manipulation tasks against representative
robot policy acceleration approaches. Across the two tasks, \method improves
task success by $25$ pp on average while reducing successful
execution time by $36.5\%$. These results demonstrate that human manipulation
tempo provides effective temporal supervision for learning faster and more
reliable robot policies.

%===============================================================================
\section{Related Work}

Faster-than-demonstration robot policies have become an important research
direction in imitation learning, aiming to improve execution efficiency beyond
the tempo of collected demonstrations. We first review existing approaches to
faster-than-demonstration robot policies and how they accelerate robot
execution. We then examine human videos as an alternative source of temporal
supervision for robot learning.

\subsection{Faster-than-Demonstration Robot Policies}

Recent work has shown that robot policies need not be constrained to the
execution tempo of their training demonstrations.
One line of work estimates the precision requirements of demonstrated motion from
policy uncertainty and applies stronger temporal compression to less
precision-sensitive segments~\cite{guo2025demospeedup}.
Another approach adapts execution speed according to motion complexity~\cite{arachchige2025sail}.
Semantic reasoning has also been used to identify trajectory regions that can
be accelerated while preserving regions requiring careful manipulation
~\cite{sanchez2026volt,kim2026espada}.
Together, these studies demonstrate that selective temporal modification can
produce faster-than-demonstration behavior without sacrificing task performance.

Beyond direct trajectory compression, execution tempo can also be learned or optimized. One approach trains a policy over speed-augmented demonstrations and subsequently refines it through reinforcement learning toward faster execution~\cite{nam2025speedaug}. Another constructs candidate future trajectories at different temporal scales and selects training targets using a prediction objective, resulting in state-dependent execution speed~\cite{hu2026autospeed}. These approaches derive their temporal adaptation from robot demonstrations, robot-side learning objectives, or robot interaction. Despite their effectiveness, existing approaches derive execution tempo primarily from robot-side information, leaving the target tempo factor without an external reference for task-appropriate manipulation speed. We therefore investigate whether human videos can provide such a reference.

\subsection{Human Videos for Robot Learning}

Human videos provide rich supervision that is difficult or expensive to obtain
from robot demonstrations alone.
Large-scale human video has been used to learn reusable visual representations
for downstream robot manipulation~\cite{nair2022visualrepresentation}.
Human behavior has also been used to infer task intent and guide robot policy
improvement through video-based human--robot correspondence
~\cite{bahl2022humanimitation}.
Object motion extracted from human videos can serve as an intermediate
representation connecting human demonstrations with robot control
~\cite{xu2024objectflow}.
These studies show that human videos can provide visual, semantic, and
object-centric motion information useful for robot learning.

More recent work has moved toward direct human-to-robot policy transfer using
egocentric demonstrations.
Egocentric human observations and 3D hand motion have been used jointly with
robot demonstrations for policy training
~\cite{kareer2025egocentricimitation}.
Cross-embodiment adaptation has also been used to explicitly align human and
robot policy representations~\cite{punamiya2025domainadaptation}, while large-scale human
pretraining and aligned human--robot training have been explored for dexterous
manipulation~\cite{zheng2026egocentricscaling}.
At a larger scale, generalist vision--language--action models have shown that
egocentric human demonstrations can be incorporated alongside robot data and
improve transfer to tasks or scenarios demonstrated only by humans, even without
an explicit human-to-robot transfer mechanism~\cite{kareer2026humanrobottransfer}.
Large collaborative egocentric datasets further support systematic study of
human-to-robot transfer at scale~\cite{punamiya2026egocentricdataset}.

These approaches demonstrate effective transfer of visual, semantic, motion,
and behavioral knowledge from human videos to robot policies.
However, these video-based approaches do not explicitly use human manipulation
tempo as a temporal reference for robot execution.
Prior work has shown that humans can directly teach robot execution speed
through physical interaction with the robot~\cite{nemec2018human}.
Here, we instead investigate whether human manipulation videos can provide
phase-wise temporal supervision without physical interaction, using phase
timing as a task-appropriate reference for robot execution tempo.

%===============================================================================
\section{Preliminaries}
\label{sec:preliminaries}

We consider action-chunk imitation learning, where a policy predicts a sequence
of future robot actions from the current observation. We then introduce
tempo-based action retiming, which forms the basis for \method.

\subsection{Imitation Learning with Action-Chunking}
\label{sec:action_chunk_il}

Let $\mathcal{D}_R=\{\tau_e^R\}_{e=1}^{N_R}$ denote a set of robot
demonstrations, where
$\tau_e^R=\{(o_{e,t},a_{e,t})\}_{t=0}^{T_e-1}$.
Given observation $o_{e,t}$, an action-chunk policy predicts $H$ future
actions~\cite{zhao2023actionchunking},
\begin{equation}
    \mathbf{A}_{e,t}
    =
    [a_{e,t},a_{e,t+1},\ldots,a_{e,t+H-1}].
    \label{eq:original_action_chunk}
\end{equation}
Standard behavior cloning therefore learns action targets at the temporal
spacing of the original robot demonstrations.

\subsection{Tempo-Based Action Retiming}
\label{sec:tempo_retiming}

The execution tempo can be increased by temporally resampling future action
targets while keeping the action-chunk length fixed
~\cite{nam2025speedaug}.
Given a tempo factor $v\geq1$, we construct the retimed target as
\begin{equation}
    \widetilde{\mathbf{A}}_{e,t}(v)
    =
    \left[
        a_e\!\left(t+(j+1)v-1\right)
    \right]_{j=0}^{H-1},
    \label{eq:retimed_action_chunk}
\end{equation}
where actions at non-integer indices are obtained by linear interpolation.
When $v=1$, the original action chunk is recovered, whereas $v>1$ selects
actions farther into the demonstrated future and thereby encourages faster
task progression.
The remaining question is how to determine an appropriate tempo $v$;
\method derives it from human manipulation timing.

\begin{figure}[t]
    \centering
    \includegraphics[width=\linewidth]{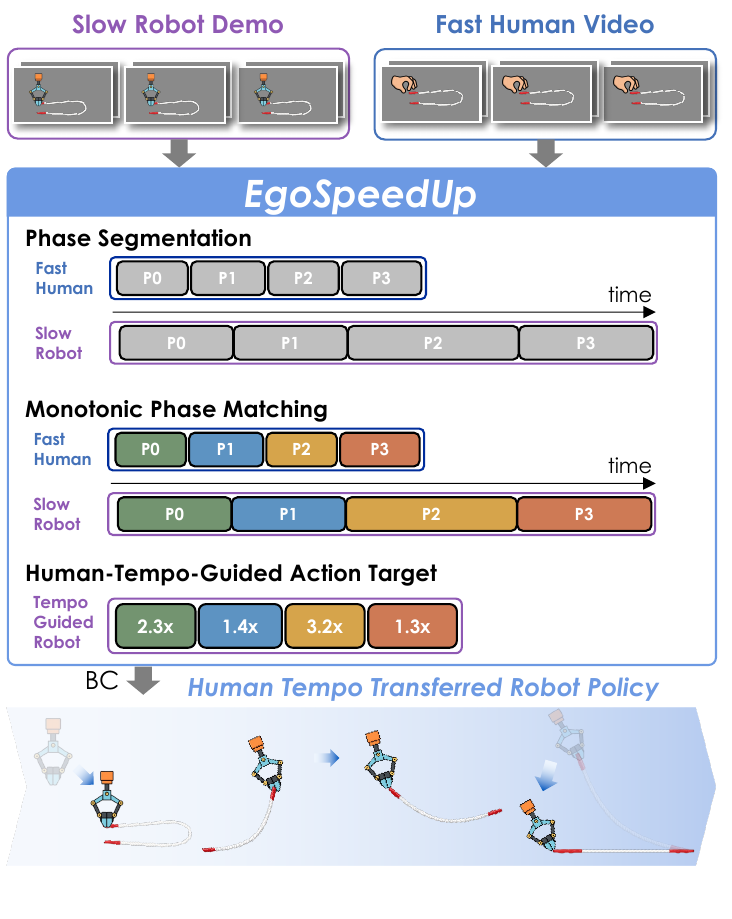}
    \caption{
    Overview of \method.
    Fast human videos and slow robot demonstrations are segmented into
    manipulation phases and aligned through monotonic phase matching.
    Corresponding phase lengths provide tempo ratios that determine phase-wise target tempos for the robot demonstrations.
    These tempos are used to retime future robot action targets, forming
    the EgoSpeedUp training data for behavior cloning.
    }
    \label{fig:method_overview}
\end{figure}
%===============================================================================
\section{EgoSpeedUp}
\label{sec:egospeedup}

\fref{fig:method_overview} illustrates the overall pipeline.
Given human and robot demonstrations of the same task, \method segments
both into ordered manipulation phases and establishes correspondence through
monotonic phase matching.
The corresponding phase lengths in steps are compared across multiple
human references to obtain a target tempo factor for each robot episode and
phase.
These human-derived tempos retime future robot action targets used for
behavior cloning.
Unlike a single task-level acceleration factor, the target tempo is estimated
separately for each episode and phase.

\subsection{Human--Robot Temporal Alignment}
\label{sec:human_robot_alignment}

Comparing total sequence lengths provides only a global tempo ratio and does not distinguish the temporal requirements of different manipulation phases.
We therefore establish correspondence at the level of task
interaction phases.
Let $\mathcal{D}_H=\{\tau_m^H\}_{m=1}^{M}$ denote the human reference
demonstrations for a given task.
We consider demonstrations that follow a shared ordered sequence of $P$
phases, such as approach, grasp, object manipulation, and release.
Semantic subtask annotations are obtained using an Embodied Chain-of-Thought (ECoT) annotation pipeline~\cite{zawalski2024embodiedcot}
and consolidated into this task-specific phase schema.
To prevent segmentation errors from obscuring the effect of tempo transfer,
we refine the semantic phase boundaries using physical interaction cues,
following prior work that segments manipulation around grasp--release events
and changes in hand--object motion~\cite{kyrarini2019assemblylearning,
kang1994motionbreakpoints,hendrich2010multisensorsegmentation}.
Robot phases are anchored to gripper events and manipulation progress, while
human phases are identified from hand--object interaction, coupled object motion,
and release.
This produces comparable temporal units without requiring equal phase
durations across demonstrations.

For robot episode $e$, let
$0=b^R_{e,0}<\cdots<b^R_{e,P}=T_e$
denote the phase boundaries in the recorded frame sequence.
Likewise,
$0=b^H_{m,0}<\cdots<b^H_{m,P}=T_m^H$
denotes the boundaries of human demonstration $m$.
The corresponding phase intervals are
\begin{equation}
    \begin{aligned}
        S^R_{e,p} &= [b^R_{e,p},b^R_{e,p+1}),\\
        S^H_{m,p} &= [b^H_{m,p},b^H_{m,p+1}),
    \end{aligned}
    \qquad p=0,\ldots,P-1.
    \label{eq:phase_intervals}
\end{equation}
Under the shared phase schema, matching associates robot phase $p$ with
human phase $p$ in each reference demonstration.
This correspondence preserves the order of task progression and constitutes
the monotonic phase matching shown in \fref{fig:method_overview}.
The original phase lengths in steps are retained for tempo estimation.

\subsection{Human-Guided Target Tempo Factor Estimation}
\label{sec:human_tempo_estimation}

Each aligned phase pair provides a relative tempo estimate by comparing the lengths of the corresponding robot and human phases in steps.
Let $\ell^R_{e,p}$ and $\ell^H_{m,p}$ denote the lengths in steps of robot and human phase $p$, respectively:
\begin{equation}
    \begin{aligned}
        \ell^R_{e,p}
        &= b^R_{e,p+1}-b^R_{e,p},\\
        \ell^H_{m,p}
        &= b^H_{m,p+1}-b^H_{m,p}.
    \end{aligned}
    \label{eq:phase_lengths}
\end{equation}
The pairwise tempo ratio is
\begin{equation}
    r_{e,p,m}
    =
    \frac{\ell^R_{e,p}}{\ell^H_{m,p}}.
    \label{eq:pairwise_tempo}
\end{equation}
A ratio greater than one indicates that the corresponding human phase
occupies fewer steps than the robot phase and therefore provides
a reference for increasing the robot target tempo.

To summarize timing across multiple human demonstrations, we average the
pairwise ratios:
\begin{equation}
    \bar r_{e,p}
    =
    \frac{1}{M}
    \sum_{m=1}^{M} r_{e,p,m},
    \qquad
    \hat v_{e,p}
    =
    \max\{1,\bar r_{e,p}\}.
    \label{eq:target_tempo}
\end{equation}
The lower bound retains the original target tempo factor when the aggregated ratio would otherwise imply slowing down the robot demonstration.
Importantly, the aggregation is performed over individual human--robot
phase-length ratios rather than over human phase lengths before division.
The resulting $\hat v_{e,p}$ is a deterministic target tempo factor for robot
episode $e$ and phase $p$.
Its episode dependence accounts for variation in robot phase lengths,
while its phase dependence retains the temporal structure provided by
the human references.

\subsection{Tempo-Guided Action Target Construction}
\label{sec:tempo_guided_targets}

For each robot training window, \method selects the target tempo factor associated
with the manipulation phase containing its starting timestep.
Let $p_e(t)$ denote this phase for timestep $t$ of episode $e$. The tempo used
for action-target construction is
\begin{equation}
    v_{e,t} = \hat v_{e,p_e(t)}.
    \label{eq:window_tempo}
\end{equation}
Using this tempo in the retiming procedure defined in
Sec.~\ref{sec:tempo_retiming}, the original action chunk
$\mathbf{A}_{e,t}$ is replaced by the retimed target
$\widetilde{\mathbf{A}}_{e,t}(v_{e,t})$ during behavior cloning.
The robot observations remain unchanged, so the human-derived phase-wise tempo
enters policy learning directly through the retimed action targets.

The policy is trained using the original robot observations and the
retimed action targets:
\begin{equation}
    \mathcal{L}_{\mathrm{EgoSpeedUp}}(\theta)
    =
    \mathbb{E}_{(e,t)\sim\mathcal{D}_R}
    \left[
        \ell_{\mathrm{BC}}
        \left(
            \theta;o_{e,t},\widetilde{\mathbf{A}}_{e,t}
        \right)
    \right].
    \label{eq:egospeedup_objective}
\end{equation}
The observation timestamps and sampling procedure remain unchanged;
human-derived timing enters the learning objective through the action targets.
In our experiments, we use a flow-based visuomotor policy
backbone~\cite{yan2025maniflow}
and retain its training loss and inference procedure.
At deployment, the learned policy predicts action chunks from robot
observations without requiring phase annotations or a tempo lookup.

%===============================================================================
\section{Experiments}
\label{sec:experiments}

We evaluate \method on real-robot manipulation tasks to address two questions:
\textbf{Q1:} Does Human-Derived Tempo Improve the
Success--Execution-Time Trade-off over Alternative Acceleration Strategies?
\textbf{Q2:} How does the effect of human-derived tempo vary
across tasks with different temporal demands?
We then examine the design choices underlying human-derived tempo estimation
and analyze the resulting tempo profiles and failure patterns.

%-------------------------------------------------------------------------------
\subsection{Experimental Setup}
\label{sec:experimental_setup}
\begin{figure*}[t]
    \centering
    \includegraphics[width=\linewidth]{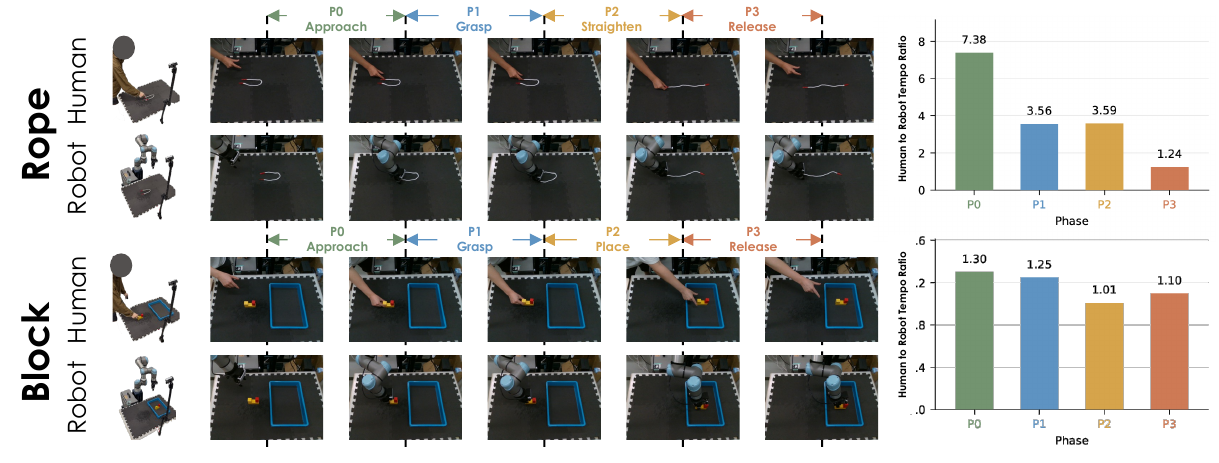}
    \caption{
    Human--robot phase correspondence and phase-wise tempo ratios for the Rope and
    Block tasks. Representative human and robot demonstrations are aligned into
    four manipulation phases ($P0$--$P3$). 
    The corresponding mean tempo ratios are computed from phase lengths and provide the temporal reference used by \method.
    }
    \label{fig:tasks_phase_correspondence}
\end{figure*}

As shown in Fig.~\ref{fig:tasks_phase_correspondence}, the real-robot setup
consists of a Universal Robots UR5e manipulator equipped with a Robotiq Hand-E parallel gripper and two Intel RealSense D435 RGB-D cameras, providing a fixed workspace view and a wrist view.
The observation consists of two consecutive RGB images from both cameras,
resized to $224\times224$, together with robot proprioception.
The seven-dimensional action consists of six joint-position targets and
one gripper command.

\subsubsection{Tasks}
We consider two manipulation tasks with different object dynamics and execution requirements: rope straightening and block placement. In \textbf{Rope}, the robot grasps a rope and manipulates it into a straight configuration, involving both grasp acquisition and dynamic manipulation of a deformable object, where rapid motion can facilitate straightening. In \textbf{Block}, the robot transports a yellow base block supporting a tower of two stacked red blocks to a target region while preserving the stability of the stacked structure, making excessive acceleration potentially detrimental to task success. These tasks therefore provide contrasting temporal demands for evaluating whether human-derived tempo adapts the degree of acceleration to the task.

\subsubsection{Human and Robot Demonstration}

For each task, we use 80 robot demonstrations collected through teleoperation
with a 3D mouse and 40 egocentric human demonstrations of the same manipulation.
The robot demonstrations cover variations in the initial end-effector and
object configurations, while the human demonstrations provide reference timing
for the corresponding manipulation phases.
All compared methods use the same robot training set. 
Human and robot demonstrations are recorded at approximately 20 Hz and
10 Hz, respectively. Target tempo factors are computed from phase lengths in recorded steps, as defined in \sref{sec:human_tempo_estimation}; the fixed sampling rates therefore
induce a constant scaling of the corresponding wall-clock duration ratios.

\subsubsection{Task-Specific Phase Segmentation}

We segment each human and robot demonstration into a shared sequence of
task-specific manipulation phases following the temporal-alignment procedure
described in Sec.~\ref{sec:human_robot_alignment}.
For Rope, we define four phases: approach, grasp, straighten, and release;
for Block, the phases are approach, grasp, place, and release, as shown in
Fig.~\ref{fig:tasks_phase_correspondence}.
The initial phase boundaries are obtained from semantic annotations generated
using ECoT~\cite{zawalski2024embodiedcot} and then refined
using task-specific physical interaction events.

For robot demonstrations, grasp and release boundaries are anchored to gripper
closing and opening events, respectively.
The onset of object manipulation is identified from the corresponding task
motion: the start of rope straightening for Rope and the start of block
transport toward the tray for Block.
These events are used to refine the initial semantic boundaries and determine
the final robot phase segmentation.

For human demonstrations, equivalent gripper events are not directly
available, so the corresponding interaction events are inferred from hand and
object observations.
2D hand keypoints are detected using WiLoR~\cite{potamias2025wilor}, and the
thumb and index fingertips are tracked in 3D using the depth stream from the
RGB-D camera described in the experimental setup.
Hand opening and closing are detected from the 3D distance between the two
fingertips, while the manipulated object is tracked using
SAM 2~\cite{ravi2025sam}.
Grasp onset is identified from hand--object proximity together with
task-specific object responses---rope deformation for Rope and sustained
proximity followed by coupled hand--object motion for Block---while the
manipulation phase begins with sustained rope straightening or block transport
toward the tray, respectively.
Release is identified from hand--object separation and hand opening; for Block,
stable placement inside the tray is additionally used to confirm the release
boundary.
These interaction events refine the initial semantic boundaries.
Fig.~\ref{fig:tasks_phase_correspondence} shows the resulting human--robot phase correspondence
and the mean phase-wise tempo ratio defined in \sref{sec:human_tempo_estimation}.

\subsubsection{Compared Methods}
\label{sec:compared_methods}
All policies use the same ManiFlow backbone~\cite{yan2025maniflow}.
See Appendix~\ref{app:implementation} for implementation details of the ManiFlow backbone.
We compare \method against the following methods:
\begin{itemize}
\item \textbf{Nominal}: trains on the original robot action targets without temporal modification.

\item \textbf{Global $2\times$ / $4\times$}: applies a fixed target tempo factor uniformly across each robot demonstration.

\item \textbf{DemoSpeedup}: applies entropy-guided demonstration acceleration, using a $2\times$ target tempo factor for segments requiring higher precision and a $4\times$ target tempo factor for segments with lower precision requirements, following the original implementation~\cite{guo2025demospeedup}.

\item \textbf{SpeedAug-prior}: trains on action targets augmented using target tempo factors from a predefined $1\times$--$3\times$ range, following the original implementation~\cite{nam2025speedaug}. We evaluate the tempo-augmented prior before reinforcement-learning fine-tuning to isolate the effect of tempo augmentation.

\item \textbf{AutoSpeed}: selects action targets using target tempo factors from a predefined $0.8\times$--$2.2\times$ range based on a robot-side prediction objective, following the original implementation~\cite{hu2026autospeed}.

\end{itemize}
We further analyze \method through ablations on global versus phase-wise target tempo factors, the number of human references, and the aggregation rule (see \sref{sec:exp:abl}).

\subsubsection{Evaluation Metrics}

We report task success rate, and the mean/median execution times over
successful episodes.
Speedup is defined as the mean successful execution time of Nominal divided by
that of the compared method within the same task.
Consequently, speedup describes successful-episode duration rather than
throughput or the cost of failed attempts.
We assess success and execution time jointly, since a short completion time
among successful episodes can coexist with a low overall success rate.

%-------------------------------------------------------------------------------
\subsection{Quantitative Results}
\label{sec:quantitative_results}

Table~\ref{tab:main_results} summarizes the results for both tasks.
We organize the comparison around two questions, separating the effect of
human-derived tempo from comparisons with alternative acceleration methods
and differences between tasks.

\begin{table*}[t]
\centering
\caption{
Main results on Rope and Block.
Each method is evaluated on 20 trials per task (two policy seeds $\times$ 10 initial poses). Total is the average over the two tasks.
Bold indicates the highest observed success rate within each task and overall.
$^{*}$ indicates that EgoSpeedUp significantly outperforms the corresponding baseline (two-sided paired $t$-tests, $p<0.05$).
}
\label{tab:main_results}
\footnotesize
\setlength{\tabcolsep}{2.7pt}
\renewcommand{\arraystretch}{1.12}
\begin{tabular}{lrrrrrrrrrr}
\toprule
 & \multicolumn{4}{c}{Rope} & \multicolumn{4}{c}{Block} & \multicolumn{2}{c}{Total} \\
\cmidrule(lr){2-5}\cmidrule(lr){6-9}\cmidrule(lr){10-11}
Method & Succ. & Mean (s) & Med. (s) & Speedup & Succ. & Mean (s) & Med. (s) & Speedup & Succ. & Speedup \\
\midrule
Nominal & 12/20 (60\%) & 14.76 & 14.31 & 1.00$\times$ & 11/20 (55\%) & 14.48 & 14.33 & 1.00$\times$ & 23/40 (57.5\%)\textsuperscript{*} & 1.00$\times$\textsuperscript{*} \\
Global 2x & 13/20 (65\%) & 8.54 & 8.39 & 1.73$\times$ & 9/20 (45\%) & 7.98 & 8.09 & 1.81$\times$ & 22/40 (55.0\%)\textsuperscript{*} & 1.77$\times$ \\
Global 4x & 11/20 (55\%) & 5.92 & 5.90 & 2.49$\times$ & 5/20 (25\%) & 4.87 & 4.83 & 2.97$\times$ & 16/40 (40.0\%)\textsuperscript{*} & 2.71$\times$ \\
DemoSpeedup & 10/20 (50\%) & 5.96 & 5.95 & 2.47$\times$ & 6/20 (30\%) & 5.29 & 5.23 & 2.74$\times$ & 16/40 (40.0\%)\textsuperscript{*} & 2.60$\times$ \\
SpeedAug-prior & 7/20 (35\%) & 8.71 & 9.00 & 1.70$\times$ & 6/20 (30\%) & 7.92 & 8.13 & 1.83$\times$ & 13/40 (32.5\%)\textsuperscript{*} & 1.76$\times$ \\
AutoSpeed & \textbf{17/20 (85\%)} & 17.16 & 16.99 & 0.86$\times$ & 10/20 (50\%) & 13.07 & 12.10 & 1.11$\times$ & 27/40 (67.5\%) & 0.97$\times$\textsuperscript{*} \\
\rowcolor{black!6}\textbf{EgoSpeedUp} & 16/20 (80\%) & 5.60 & 5.66 & 2.63$\times$ & \textbf{17/20 (85\%)} & 12.89 & 12.82 & 1.12$\times$ & \textbf{33/40 (82.5\%)} & 1.58$\times$ \\
\bottomrule
\end{tabular}
\end{table*}

\subsubsection{Q1: Does Human-Derived Tempo Improve the
Success--Execution-Time Trade-off over Alternative Acceleration Strategies?}

We first examine the aggregate performance across both tasks.
\method achieves the highest pooled success rate of $82.5\%$ while providing a
$1.58\times$ overall speedup relative to Nominal.
In comparison, methods that achieve larger speedups show substantially lower
pooled success: Global $4\times$ and DemoSpeedup reach $2.71\times$ and
$2.60\times$ speedup, but only $40.0\%$ success, while SpeedAug achieves
$1.76\times$ speedup with $32.5\%$ success.
Conversely, AutoSpeed attains the second-highest pooled success rate
($67.5\%$) but does not improve execution speed over Nominal
($0.97\times$).

Overall, these results indicate that human-derived temporal guidance provides a
favorable balance between task completion and execution speed rather than
simply maximizing acceleration.
This suggests that task-appropriate manipulation tempo encoded in human
demonstrations can contribute not only to faster execution but also to reliable
task completion.

\subsubsection{Q2: How Does the Effect of Human-Derived Tempo Vary
Across Tasks with Different Temporal Demands?}

The effect of human-derived tempo differs substantially between the two tasks.
On Rope, \method primarily improves execution efficiency: it reduces mean
successful execution time from $14.76\,\mathrm{s}$ to $5.60\,\mathrm{s}$,
corresponding to a $2.63\times$ speedup, while also increasing success from
$60\%$ to $80\%$.
AutoSpeed achieves slightly higher success ($85\%$), but requires
$17.16\,\mathrm{s}$ on average, whereas Global $4\times$ and DemoSpeedup reach
similar execution times to \method at considerably lower success rates of
$55\%$ and $50\%$.
Thus, the main effect of human-derived tempo on Rope is aggressive acceleration
without the reliability loss observed for the faster fixed and
demonstration-derived alternatives.

On Block, the effect is different.
\method increases success from $55\%$ to $85\%$, while mean successful execution
time decreases only moderately from $14.48\,\mathrm{s}$ to $12.89\,\mathrm{s}$
($1.12\times$ speedup).
More aggressive acceleration achieves much shorter successful execution times,
but substantially reduces success: Global $4\times$ and DemoSpeedup achieve
$25\%$ and $30\%$ success, respectively.
Here, the primary benefit of human-derived tempo is therefore improved task
completion rather than large execution-time reduction.

These contrasting outcomes are consistent with the task-dependent tempo
profiles in Fig.~\ref{fig:tempo_comparison}.
Rope receives substantially stronger phase-wise acceleration, whereas Block is
assigned a more moderate tempo profile.
The results suggest that human temporal supervision adapts the degree of
acceleration to the manipulation demands of each task.

%-------------------------------------------------------------------------------
\subsection{Analysis of Human-Derived Tempo}
\label{sec:human_tempo_analysis}

We next analyze how human-derived tempo is constructed and how it affects robot
execution.
The design analysis examines the temporal scope, number of human references,
and aggregation rule used to estimate the target tempo factor.
The behavioral analysis compares the resulting tempo profiles across methods
and examines the corresponding failure counts.

\begin{table}[t]
\centering
\caption{
Design analysis relative to the default \method configuration
(phase-wise tempo, $M=40$ human references, and pairwise mean aggregation).
The first row reports the absolute performance of \method.
All subsequent rows report changes relative to this reference.
}
\label{tab:design_analysis}
\footnotesize
\setlength{\tabcolsep}{5.5pt}
\renewcommand{\arraystretch}{1.12}
\begin{tabular}{lrr}
\toprule
Method
& Succ. $\uparrow$ & Time (s) $\downarrow$ \\
\midrule
\rowcolor{black!6}
\textbf{EgoSpeedUp}
& \textbf{82.5\%} & \textbf{9.25} \\
\midrule
EgoSpeedUp w/ Global Tempo
& $-17.5$ pp & $+1.76$ \\
EgoSpeedUp w/ $M=5$
& $-22.5$ pp & $+0.16$ \\
EgoSpeedUp w/ $M=10$
& $-17.5$ pp & $+0.25$ \\
EgoSpeedUp w/ Median Aggregation
& $-20$ pp & $+0.18$ \\
\bottomrule
\end{tabular}
\end{table}

\subsubsection{Design Analysis}
\label{sec:exp:abl}
Table~\ref{tab:design_analysis} evaluates three design choices associated with
the target tempo estimation in Eq.~\eqref{eq:target_tempo}: phase-wise versus
global estimation, the number of human references $M$, and pairwise-ratio
aggregation (mean versus median).

\paragraph{Phase-Wise vs. Global Tempo}
We compare the default phase-wise target tempo factors with a single global tempo factor computed for each robot episode as the ratio of its total duration to the mean total duration of the human demonstrations, which is applied uniformly across all phases.
Replacing phase-wise tempo with global tempo reduces overall success by
$17.5$ pp and increases mean successful execution time by
$1.76\,\mathrm{s}$.
These results support phase-wise estimation of human tempo as a better
approximation of task-appropriate manipulation tempo than a single global
estimate.

\paragraph{Effect of the Number of Human Demonstrations}
We compare $M\in\{5,10,40\}$ human references with all other settings fixed.
Relative to $M=40$, using $M=5$ and $M=10$ lowers overall success by
$22.5$ and $17.5$ pp, respectively, while increasing mean successful
execution time by only $0.16,\mathrm{s}$ and $0.25,\mathrm{s}$.
Phase durations can vary depending on the initial object configuration, making
the estimated target tempo factors sensitive to the diversity of the human
reference set.
With fewer human references, the estimated phase-wise target tempo factors
may be biased toward particular initial object configurations, making the
resulting temporal supervision less representative of other configurations.
A larger reference set can better capture this variation, yielding more
representative target tempo factors across different initial object
configurations and thereby improving task success.

\paragraph{Aggregation Strategy}
We use the pairwise mean as the default aggregation to estimate the average
relative tempo across the human reference set.
Unlike the mean, median aggregation does not reflect the magnitude of tempo
ratios away from the center of the distribution, which may underrepresent meaningful temporal variation across different initial object configurations.
Consistent with this interpretation, replacing the pairwise mean with the
median reduces overall success by $20.0$ pp while increasing
mean successful execution time by only $0.18\,\mathrm{s}$.

\subsubsection{Behavioral Analysis}

\begin{figure}[t]
    \centering
    \includegraphics[width=\linewidth]{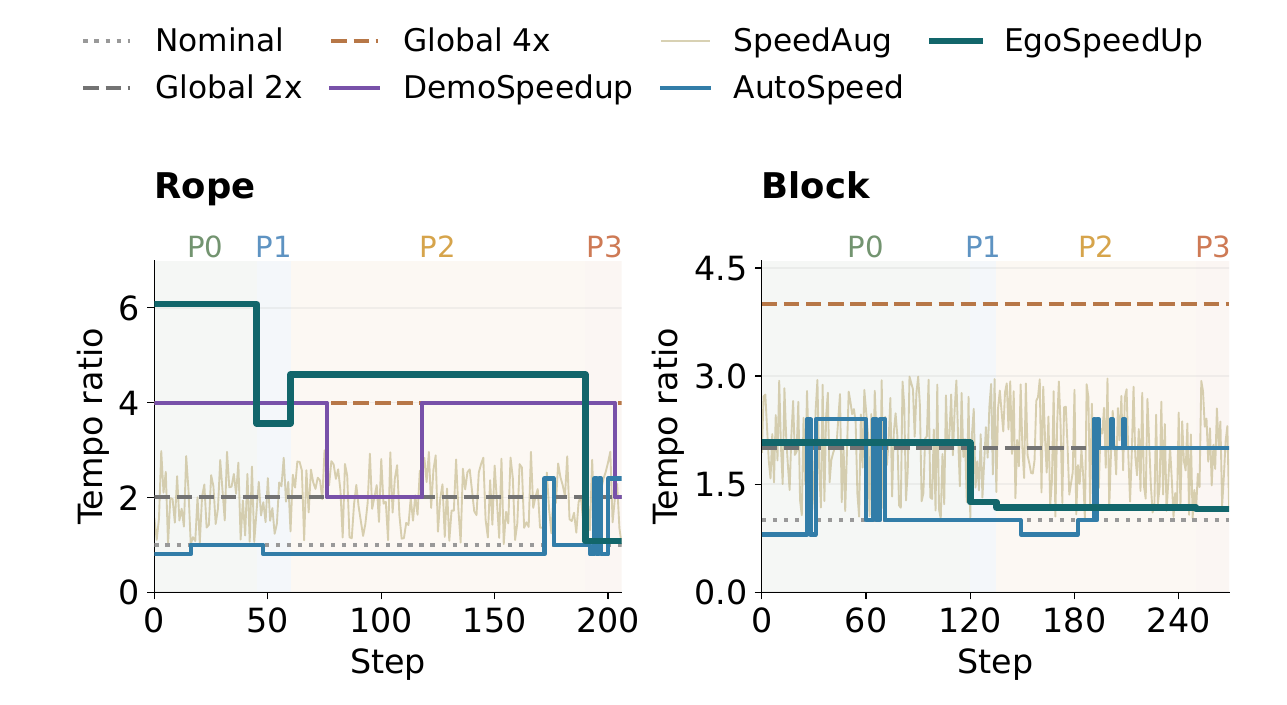}
    \caption{
    Target tempo profiles for representative Rope and Block demonstrations.
    The horizontal axis denotes the original demonstration step, and the curves
    show the target tempo factor assigned to each training window.
    Global baselines use fixed tempo factors, whereas \method assigns
    piecewise-constant target tempo factors derived from aligned human manipulation
    phases.
    The plotted values are training targets, not measured rollout speedups.
    }
    \label{fig:tempo_comparison}
\end{figure}

\paragraph{Phase-Wise Tempo Profiles}
Fig.~\ref{fig:tempo_comparison} compares the target tempo factors produced by the
evaluated methods.
The baseline profiles reflect their predefined tempo settings described in
Sec.~\ref{sec:compared_methods}.
Global $2\times$ and $4\times$ remain fixed at their prescribed factors,
DemoSpeedup switches between $2\times$ and $4\times$, SpeedAug varies within
its predefined $1\times$--$3\times$ range, and AutoSpeed selects targets within
its predefined $0.8\times$--$2.2\times$ range.

In contrast, \method does not prespecify either discrete tempo factors or a
bounded tempo range, but derives its target tempo factors directly from human
manipulation videos. For Rope, \method assigns approximately $6\times$ tempo in $P0$,
$3.6\times$ in $P1$, and $4.6\times$ in $P2$, before decreasing to
approximately $1\times$ in $P3$.
For Block, the target tempo factor is substantially more moderate, decreasing from
approximately $2.1\times$ in $P0$ to about $1.2\times$ in the later phases.
Thus, \method selects different target tempo factors across phases and tasks according
to the corresponding human manipulation timing.

These results highlight an important distinction from the compared acceleration
baselines, which prespecify either discrete tempo factors or a bounded range of
tempo adjustment.
Rather than determining in advance how much the robot should be accelerated,
\method derives target tempo factors directly from human manipulation timing.
This removes the need to manually define an acceleration range and provides flexibility to adapt the target tempo factor across new tasks and objects.

\begin{figure}[t]
    \centering
    \includegraphics[width=\linewidth]{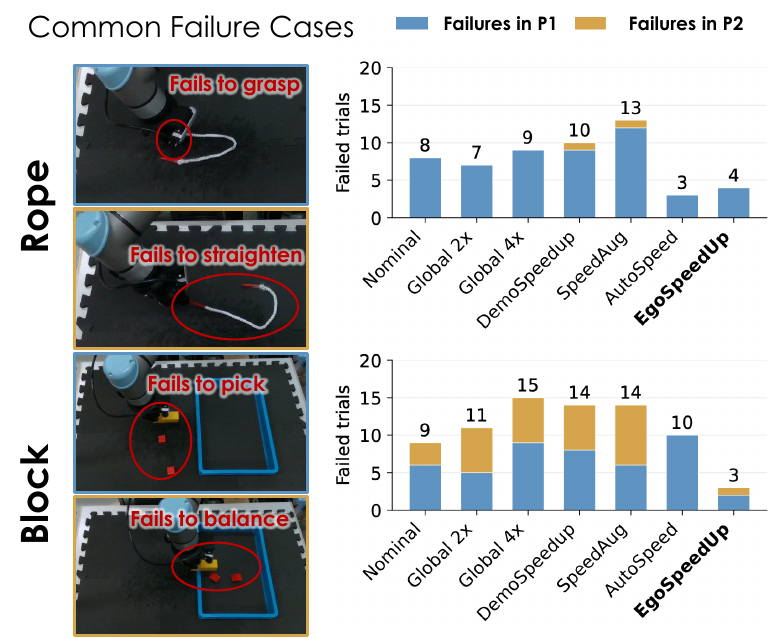}
    \caption{
    Representative failure cases and phase-associated failure counts for Rope
    and Block.
    Rope failures are categorized as grasping and straightening, and Block
    failures as picking and unstable placement.
    Stacked bars report failed trials among 20 attempts per method and task;
    numbers above the bars indicate total failures.
    Failure counts are not normalized by the number of trials that reached each phase.
    }
    \label{fig:failure_cases}
\end{figure}

\paragraph{Failure Case Analysis}
Fig.~\ref{fig:failure_cases} reveals distinct failure patterns across the two
tasks.

Most failures on Rope occur during grasping.
\method records four grasping failures and no straightening failures, compared
with eight grasping failures for Nominal.
DemoSpeedup and SpeedAug record nine and twelve grasping failures,
respectively, together with one straightening failure each.
Thus, the higher Rope success rate of \method is primarily associated with
fewer grasping failures.
AutoSpeed records one fewer failure than \method, but at a substantially longer
execution time.

On Block, \method reduces failures in both object acquisition and final
placement.
It records two picking failures and one unstable-placement failure, compared
with six and three, respectively, for Nominal.
More aggressive baselines exhibit substantially larger failure counts,
consistent with their lower success rates in
Table~\ref{tab:main_results}.
This suggests that the higher Block success rate of \method is associated with
fewer failures across multiple stages rather than improvement at a single
stage.

%%%%%%%%%%%%%%%%%%%%%%%%%%%%%%%%%%%%%%%%%%%%%%%%%%%%%%%%%%%%%%%%%%%%%%%%%%%%%%%%
\section{Discussion}

EgoSpeedUp shows that human manipulation tempo can serve as phase-wise temporal supervision for accelerating robot policies. Despite this, EgoSpeedUp has several limitations. First, it assumes that human and robot demonstrations can be aligned through a shared ordered sequence of manipulation phases; repeated, branching, or weakly structured tasks may require more flexible correspondence mechanisms, such as trajectory-level alignment across human and robot demonstrations~\cite{liu2025immimic}. Second, EgoSpeedUp assigns a single deterministic target tempo factor to each aligned phase, which cannot capture within-phase tempo variation or state-dependent changes in execution difficulty. Combining human-derived tempo with uncertainty- or state-dependent adaptation, as explored in entropy-guided acceleration~\cite{guo2025demospeedup} and stage-adaptive speed selection~\cite{hu2026autospeed}, could enable finer-grained tempo adaptation while preserving human tempo as an external reference.

%%%%%%%%%%%%%%%%%%%%%%%%%%%%%%%%%%%%%%%%%%%%%%%%%%%%%%%%%%%%%%%%%%%%%%%%%%%%%%%%
\section{Conclusion}

We presented EgoSpeedUp, a framework that transfers human manipulation tempo to robot policies through phase-wise temporal supervision. By aligning corresponding manipulation phases, estimating human-derived target tempo factors, and retiming robot action targets accordingly, EgoSpeedUp enables standard behavior cloning to learn robot policies with task-appropriate execution tempo. Across two real-world manipulation tasks, EgoSpeedUp achieved a favorable trade-off between success rate and execution time compared with state-of-the-art methods, providing strong acceleration on Rope while substantially improving task success on Block. These results demonstrate that human manipulation tempo can serve as an effective external temporal reference for learning faster and more reliable robot policies.

%%%%%%%%%%%%%%%%%%%%%%%%%%%%%%%%%%%%%%%%%%%%%%%%%%%%%%%%%%%%%%%%%%%%%%%%%%%%%%%%

\section{Acknowledgment}
This work was supported by JST CREST, Japan, Grant Number JPMJCR2553, under the research project ``MORAL: Morphoception-Oriented Reasoning and Action with Language.''

%===============================================================================
\appendices
\section{Implementation Details}
\label{app:implementation}

The compared methods are described in Sec.~\ref{sec:compared_methods}.
Table~\ref{tab:hyperparameters} reports the common ManiFlow configuration used
throughout the experiments.

\begin{table}[t]
\centering
\caption{Common hyperparameters for all evaluated ManiFlow-based methods.}
\label{tab:hyperparameters}
\scriptsize
\setlength{\tabcolsep}{3.5pt}
\renewcommand{\arraystretch}{1.05}
\begin{tabular}{@{}lrlr@{}}
\toprule
Hyperparameter & Value & Hyperparameter & Value \\
\midrule
Learning rate & $10^{-4}\!\rightarrow\!10^{-5}$ & Policy backbone & DiTX \\
AdamW betas & $(0.9,\,0.95)$ & Transformer layers & 12 \\
Weight decay & $10^{-3}$ & Hidden dimension & 768 \\
Batch size & 64 & Attention heads & 8 \\
\midrule
Visual encoder & CLIP ViT-B/16 & Flow inference steps & 10 \\
Observation steps & 2 & Action horizon & 16 \\
\bottomrule
\end{tabular}
\end{table}

\bibliographystyle{ieeetr}
\bibliography{reference}

\end{document}